\documentclass[runningheads]{llncs}
\usepackage{graphicx}
\usepackage{multirow}
\usepackage{booktabs}

\usepackage[colorinlistoftodos,prependcaption]{todonotes}
\setuptodonotes{inline, backgroundcolor=white, bordercolor=white, textcolor=red}

\begin{document}
\title{Indoor Navigation Assistance for Visually Impaired People via Dynamic SLAM and Panoptic Segmentation with an RGB-D Sensor}
\titlerunning{Navigation Assistance for the Visually Impaired}


\author{Wenyan Ou \and
Jiaming Zhang \and
Kunyu Peng \and
Kailun Yang\inst{*} \and\\
Gerhard Jaworek \and
Karin M\"uller \and 
Rainer Stiefelhagen
\vskip-1ex
}
\authorrunning{W. Ou \textit{et al.}}
\institute{Karlsruhe Institute of Technology, 76131 Karlsruhe, Germany\\
\email{Correspondence: kailun.yang@kit.edu}
}

\maketitle              
\begin{abstract}
\vskip-5ex
Exploring an unfamiliar indoor environment and avoiding obstacles is challenging for visually impaired people. Currently, several approaches achieve the avoidance of static obstacles based on the mapping of indoor scenes. To solve the issue of distinguishing dynamic obstacles, we propose an assistive system with an RGB-D sensor to detect dynamic information of a scene. Once the system captures an image, panoptic segmentation is performed to obtain the prior dynamic object information. With sparse feature points extracted from images and the depth information, poses of the user can be estimated. After the ego-motion estimation,  the dynamic object can be identified and tracked. Then, poses and speed of tracked dynamic objects can be estimated, which are passed to the users through acoustic feedback.


\end{abstract}
\section{Introduction}
Human perception of the environment often takes an eye-based approach, which makes vision an indispensable part of daily life.
People with visual impairments usually have very limited or no access to this channel.
It is known that they mainly rely on information from other modalities to gain perception of the surroundings, the most important is hearing. Therefore, once the environment of visually impaired people is too noisy, their perception in environments with dynamic objects will be deviated, resulting in collisions with obstacles and even injuries, which greatly affects their daily life.

Besides, visually impaired people find it hard to maintain proper social distances from others during the Covid-19 pandemic~\cite{martinez2020helping}.
Some assistance systems tackle this issue through Simultaneous Localization And Mapping (SLAM) and deep learning approaches~\cite{liu2021hida,zhang2021trans4trans}, to provide accurate guidance to visually impaired people, but they are less effective in highly dynamic scenarios.
To address this problem, we propose a system to help people with visual impairments perceive dynamic objects in indoor environments and understand their motion.

Recently, research on SLAM has gradually shifted from traditional static environments to more diverse dynamic environments, which are closer to reality. 
According to the further processing of dynamic objects, current solutions can be divided into two categories.
Some works directly discard the dynamic information as outliers~\cite{DynaSLAM,DS-SLAM}.
Other works maintain them and jointly optimize the static map, pose of the camera, and dynamic objects in the scene~\cite{DynaSLAM2,VDOSLAM}.

\begin{figure}[!t]
\includegraphics[width=\textwidth]{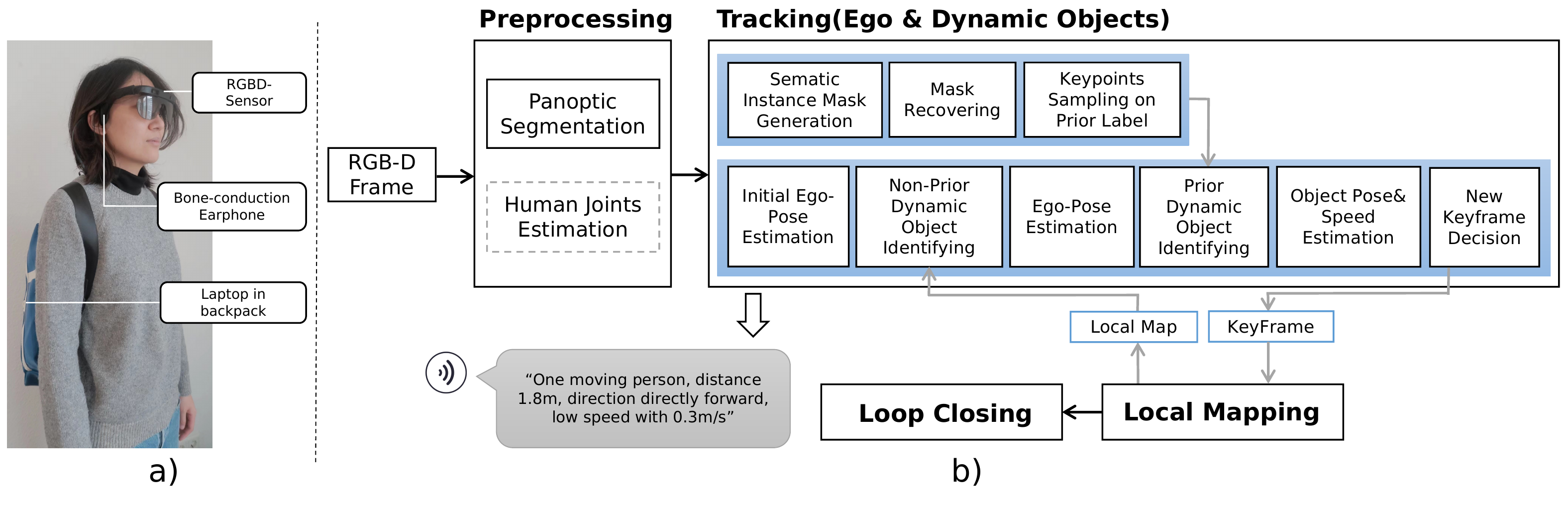}
\vskip-3ex
\caption{(a) Our wearable devices designed for assisting people with visual impairments; (b) The proposed algorithm pipeline for navigation assistance.} \label{fig1}
\vskip-3ex
\end{figure}

In this work, we propose a novel indoor assistance system for visually impaired people while dealing with dynamic environments, of which the main structure is shown in Fig.~\ref{fig1}.
We develop a wearable assistance system based on an RGB-D sensor, which estimates the user's ego-pose, together with a static feature point map.
The dynamic objects can be identified by the proposed system, and the average depth information to the user can be obtained.
When the dynamic object belongs to a prior class, \textit{e.g.}, \emph{people}, it can also be tracked between frames.
Moreover, the linear velocity of prior dynamic objects can be estimated and transmitted through the bone-conduction headphones located on our smart glasses system (see Fig.~\ref{fig1}(a)) to the user.

In summary, our main contributions are: a wearable assistant system with an RGB-D sensor is proposed, and it can be used to achieve localization and mapping and help visually impaired people detect dynamic objects in indoor scenes and obtain corresponding motion cues.

\section{Related Work}
\textbf{SLAM in a dynamic environment.}
In the past years, many visual SLAM systems were proposed and have a satisfactory performance, such as ORB-SLAM~\cite{orbslam2}, DVO-SLAM~\cite{DenseVOSLAM}, \textit{etc}.
However, these visual SLAM systems are based on the assumption of a static- or a slightly dynamic environment.
If there are highly dynamic objects in the environment - which is more closed to real-life scenarios - pose estimation and mapping may lead to poor results.
Some approaches deal with dynamic objects in a pure geometry-based way~\cite{DSLAM,DenseVOwithDD}.
Other works leverage both deep learning and geometry-based methods to eliminate the negative effects of dynamic objects~\cite{DynaSLAM,DS-SLAM,poseFusion}.
Recently, some works~\cite{DynaSLAM2,VDOSLAM} tackle this issue by tracking dynamic objects instead of removing them.

\noindent\textbf{SLAM in assistance systems.}
As for assistive applications for visual impaired people, SLAM is often used to achieve positioning of users and obstacle detection~\cite{ISANA,SmartCane}. For navigational assistive systems, obstacle detection is desired to provide more detailed information to help avoid collisions and understand the surroundings. Therefore, semantic information is also incorporated into a SLAM system to achieve semantic path- and destination finding~\cite{SemanticMap}. Besides, a prior map with semantic information can be established by SLAM in advance for navigation systems in indoor environments~\cite{priorMap,liu2021hida} and can be later used for global path planning.
Differing from these existing works, our work considers the localization and mapping in challenging dynamic environments and combines feature descriptors matching in ego-motion estimation and optical flow tracking in dynamic object motion estimation to guarantee valid tracking.
Besides, our system can help seeing impaired people avoid collisions with diverse moving obstacles.

\section{Method}
Fig.~\ref{fig1}~(a) shows our assistant system consisting of a pair of smart glasses and a lightweight laptop.
First, the surrounding environment of visual impaired people is captured by the aforementioned RGB-D sensor attached to the glasses in the wearable device.
Then, panoptic segmentation of the RGB image is executed online on a laptop with a processor, and optionally human joint keypoint estimation will be performed.
After the pre-processing step, the RGB-D image and the segmentation mask will be passed into the tracking module and further processed in the local mapping and loop closing modules.
Focusing on the perspective of human-computer interaction, 
the information about the surroundings are delivered to the users, through the bone-conduction earphones on the glasses.

\subsection{Preprocessing}
\begin{figure}[!t]
\includegraphics[width=\textwidth]{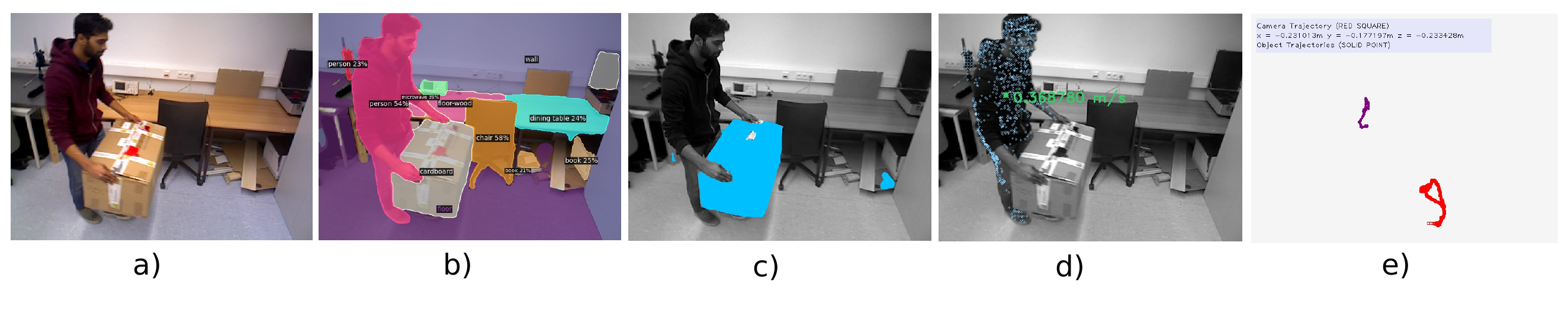}
\vskip-4ex
\caption{(a) RGB image; (b) Panoptic segmentation result; (c) Non-prior dynamic object; (d) Tracked points and the estimated speed of the prior dynamic person; (e) Top-view trajectory of the user and the moving person.} \label{fig2}
\vskip-3ex
\end{figure}

PanopticFCN~\cite{PanopticFCN} is leveraged using the obtained RGB image as input. The output of panoptic segmentation consists of semantic- and instance-based masks, as visualized in Fig.~\ref{fig2}(b).
Since people tend to move dynamically in real-life indoor scenes, the annotation of \emph{people} is set as prior dynamic.
If a more accurate speed estimation of dynamic moving people is expected, the human joint keypoints' coordinates detected by OpenPose~\cite{OpenPose} can be generated as well.
We select the joint keypoints on shoulder and middle hip, constraining the sampled points on the trunk body. So the points are more likely to be located on the slightly deformed parts of the body, enabling stable motion estimation.

\subsection{Identifying dynamic objects}
Dynamic objects that can be addressed by our system cover two categories, \textit{i.e.}, prior dynamic objects like \emph{people} or \emph{pets},
and non-prior moving objects, \textit{e.g.}, a carton passively moved by people as shown in Fig.~\ref{fig2}(c).
On the one hand, the non-prior dynamic object will be identified after the initial ego-pose estimation through the depth difference method, which is similar to~\cite{DynaSLAM}.
The main difference between our approach and \cite{DynaSLAM} is that the instance mask is directly leveraged to calculate the percentage of dynamic points in an object and identify if it is moving or not, rather than using region growing on depth images.
On the other hand, 3D scene flow of sampled points on prior objects in consecutive frames is utilized to verify if a prior object is dynamic or not.
When the magnitude of scene flow of a point is larger than a threshold, \textit{i.e.}, $0.02$ used in our work, the point will be regarded as dynamic.
The object is regarded as a dynamic object if the percentage of dynamic points in the prior object is over a threshold ($30\%$).
Considering computational costs, only dynamic prior objects will be tracked and expected to be described with speed.
The dynamic non-prior and static prior objects are feedback with the direction and averaged depth from the user.

\subsection{Ego-motion estimation}
We follow the same pipeline proposed by~\cite{orbslam2} to achieve ego-motion estimation.
In our work, we add the step of ``Non-Prior Dynamic Object Identifying'' before tracking the local map.
As for the mapping step, merely static points are considered as map points.
Thanks to the excellent feature-based visual SLAM framework, we can estimate the ego-pose robustly and accurately, which is essential to the following dynamic object tracking.

\subsection{Dynamic objects pose \& speed estimation}

We leverage a similar methodology as in~\cite{VDOSLAM} for the pose and speed estimation of the dynamic objects. Preparation procedure is firstly executed on the other thread while solving the ego-motion prediction. The potential dynamic objects with prior labels are numbered in the obtained panoptic segmentation mask respectively, while all the other pixels are annotated as $0$. Then, the DIS optical flow~\cite{DIS} is used to find corresponding keypoints in the current frame. With this dense optical flow method, if a potential prior dynamic object appears in the mask of the last frame but fails to be segmented in the current frame, the object keypoints on the last frame can be tracked and recovered in the current frame. Those dynamic objects tracked by optical flow, as presented in Fig.~\ref{fig2}(d), can be assigned unique track indices over time. To guarantee a sufficient tracking number of dynamic points, we sample every five points within an object mask. When the number of tracked points within a dynamic object decreases below the threshold, those sampled candidate points can be supplied.

After obtaining the ego-pose of current frame, the scene flow of points can be calculated. Therefore, the prior dynamic but actually remaining-static objects can be filtered according to the magnitude of scene flow. As for the initial pose estimation of dynamic objects, a better result with more inliers is selected from the EPnP method~\cite{EPnP} or using the previous motion. Finally, further pose optimization and speed calculation as in~\cite{VDOSLAM} are used to obtain the final result.

\section{Evaluation}
We test our system on some sequences of public indoor TUM RGB-D dataset~\cite{tumDataset} and Bonn RGB-D dataset~\cite{bonnDataset}. We select sequences with dynamic objects, including seated- and several walking people and \textit{etc.}, to simulate real-life scenarios in the office or different rooms that are important for visually impaired people.

\noindent\textbf{Quantitative results of pose estimation.}
The experiments of evaluating the ego-motion are carried out. We choose RMSE of absolute trajectory- and relative pose error as error metrics, which indicates robustness of the system, as proposed by~\cite{tumDataset}.
Table~\ref{tab1} presents the comparison between our system and the baseline framework ORB-SLAM2~\cite{orbslam2} as well as the ORB-SLAM2 with prior semantic information.
Our system shows better effectiveness in most cases, which verifies the superiority of our approach for highly dynamic indoor scenes.
Since the dynamic objects in the sequences of TUM RGB-D dataset are mainly with the prior label `people’, so the difference between the latter two methods is small.
For the slightly dynamic scenes like fr3/sitting rpy, ORB-SLAM2 has better performance, since it has more valid keypoints located on the people.
However, the ATE of the latter two methods are relatively small.
For the sequence crowed with three very fast moving and rotating people, the second method show a better result. But this method only maintains efficiency for sequences with prior dynamic objects. For the challenging sequences like moving nonobstructing box and moving obstructing box, our system shows a robust performance. 

\begin{table}[h]
\vskip-2ex
\scriptsize
\begin{center}
\caption{Ego-motion comparison on the TUM and Bonn datasets. The top five sequences are from TUM, whereas the bottom six sequences are from Bonn. unit: ATE (m), $RPE_t$ (m/frame), $RPE_r$ (degree/frame). ORBv2: ORB-SLAM2~\cite{orbslam2}.}\label{tab1}
\vskip-3ex
\setlength{\tabcolsep}{0.75mm}{
\begin{tabular}{c|ccc|ccc|ccc}
\toprule
\multirow{2}{*}{Sequences} & \multicolumn{3}{c|}{ORBv2 (RGB-D)} & \multicolumn{3}{c|}{ORBv2 with semantic} & \multicolumn{3}{c}{Our system} \\
& $ATE$ & $RPE_t$  & $RPE_r$ & $ATE$ & $RPE_t$ & $RPE_r$ & $ATE$ & $RPE_t$  & $RPE_r$\\ 
\midrule
fr3/walking\_static   &0.271
    &0.026
    &0.515
    &$\mathbf{0.006}$
    &0.006
    &0.174
    &$\mathbf{0.008}$
    &0.006
    &0.175
    \\
fr3/walking\_xyz   &0.786
    &0.028
    &0.652
    &$\mathbf{0.016}$
    &0.012
    &0.392
    &$\mathbf{0.015}$
    &0.012
    &0.388
    \\
fr3/walking\_halfsphere   &0.481
    &0.024
    &0.603
    &$\mathbf{0.026}$
    &0.015
    &0.434
    &$\mathbf{0.028}$
    &0.013
    &0.417
    \\
fr3/walking\_rpy   &0.840
    &0.036
    &0.760 
    &$\mathbf{0.036}$    
    &0.023
    &0.535
    &$\mathbf{0.033}$
    &0.021
    &0.494
    \\
fr3/sitting\_rpy   &$\mathbf{0.019}$
    &0.013
    &0.416
    &0.031
    &0.017
    &0.450 
    &0.035
    &0.018
    &0.450
      \\ \hline
crowd   &1.179
    &0.132
    &3.191
    &$\mathbf{0.022}$
    &0.014
    &0.654
    &0.035
    &0.014
    &0.634

      \\
moving\_nonobstructing\_box   &0.383
    &0.023
    &0.931
    &0.084
    &0.028
    &0.986
    &$\mathbf{0.037}$
    &0.023
    &0.735
    \\
moving\_obstructing\_box   &0.471
    &0.022

    &1.139
    &0.342
    &0.017
    &1.082
    &$\mathbf{0.196}$
    &0.045
    &1.616
    \\
person\_tracking   &0.632
    &0.030
    &1.581

    &$\mathbf{0.041}$
    &0.021    &1.513
    &$\mathbf{0.037}$
    &0.021
    &1.507

    \\
person\_tracking2   &0.988
    &0.037
    &1.440 
    &$\mathbf{0.048}$
    &0.018
    &1.266
    &$\mathbf{0.042}$
    &0.018
    &1.267

    \\
removing\_nonobstructing\_box   &$\mathbf{0.019}$

    &0.016
    &0.892
    &$\mathbf{0.015}$
    &0.015
   &0.886
   &$\mathbf{0.016}$
   &0.015
   &0.886
     \\
\bottomrule 
\end{tabular}}
\end{center}
\vskip-4ex
\end{table}

\begin{figure}[!t]
\includegraphics[width=\textwidth]{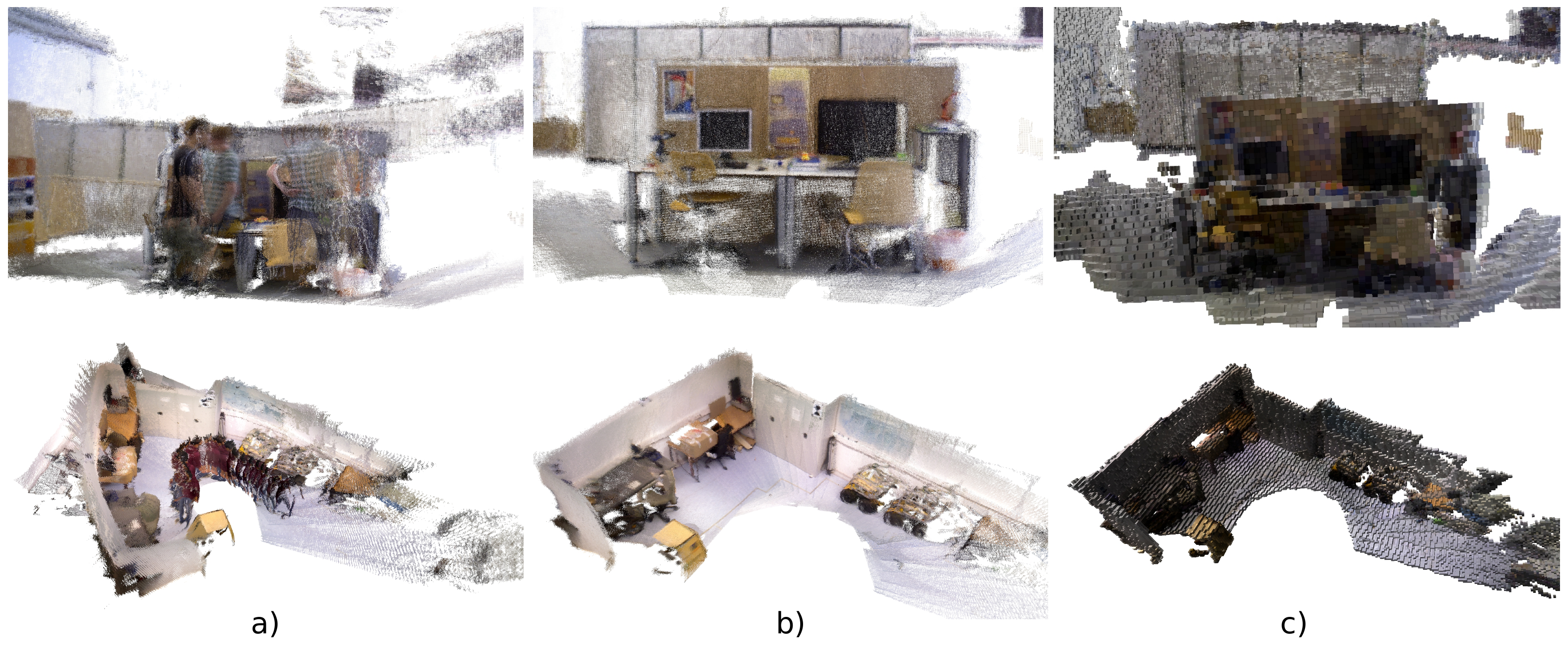}
\vskip-3ex
\caption{Generated dense point map and octree map. (a) original dense
point cloud map generated by ORB-SLAM2; (b) dense point cloud map generated by our system; (c) octree map based on our dense point cloud map, without filtering ground information.} \label{fig:map}
\vskip-4ex
\end{figure}

\noindent\textbf{Qualitative results of dynamic object estimation.}
Since the ground-truth information of the dynamic objects are not provided, a qualitative analysis of the estimation is presented in this section. We compute dynamic objects' speed as shown in Fig.~\ref{fig2}(d).
In the example scenario, the moving direction and speed of an on-coming or passing person are to be detected by the system and delivered to the user, so that they can react in time and avoid a collision. 

To investigate the map built for further navigation tasks, we evaluate the mapping capacity of our system. Since the sparse point cloud map generated by ORB-SLAM2 can not be easily used in the practical application, we generate a dense point cloud map offline after obtaining the keyframe poses for the whole sequence. In Fig.~\ref{fig:map}, we have visualized the dense point cloud with or without the effect of dynamic objects. Moreover, we also generate octree maps based on the correct dense point cloud map, which are suitable for assistive functions.
The overlapping prior dynamic objects (\emph{person}) in the point cloud are removed and the valid map is generated.
Here, one should note that during the whole sequence, the point cloud of the non-prior dynamic objects, such like the \emph{chairs} in the left sequence, still remain two versions, \textit{i.e.}, the position before it was moved by the person and the position when it became static again. Whether maintaining or deleting this kind of point cloud, a more complex strategy should be considered. For long-term SLAM systems, the map will be locally updated at a certain time. 

\noindent\textbf{Runtime analysis.}
We test the average computational time of the system. The time excludes the part of panoptic segmentation and joint keypoints extraction, as this part totally depends on the GPU type and the selection of a certain neural network. And the time cost of our system is highly related to the number of tracked dynamic objects. For all the sequences, we achieve an average speed at $4{\sim}7$ FPS on an i5-10210U CPU, which is reasonable for indoor navigation.

\begin{figure}[!t]
\includegraphics[width=\textwidth]{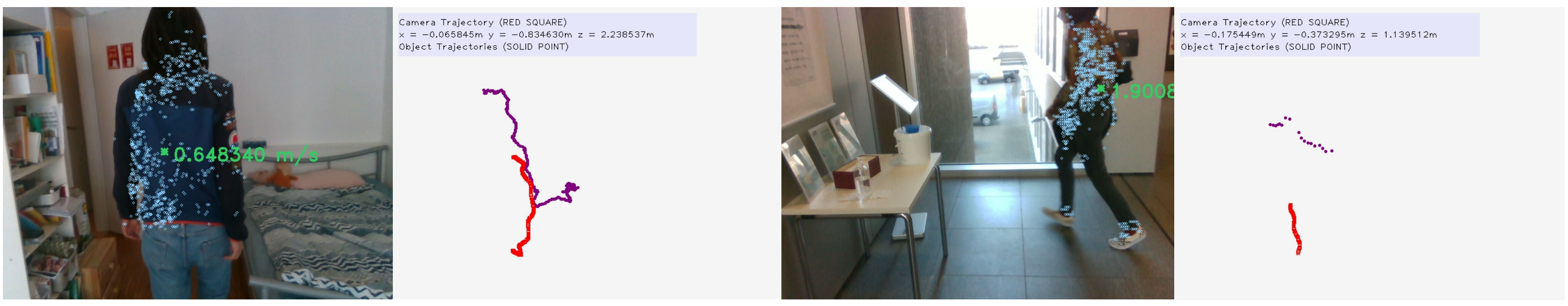}
\vskip-3ex
\caption{Speed and trajectory estimation in real-life scenarios. The possible output of first scenario: One person on front side, low speed, $1.2$ meter distance; The possible output of second scenario: One person on right side, high speed, $1.6$ meter distance. } \label{fig:real-life}
\vskip-4ex
\end{figure}

\noindent\textbf{Real-life scenarios.}
In addition to the function of localization and mapping in dynamic environments, we also try to explore the application of dynamic information in the assistive system for the visually impaired. We collected some sequences in real-life scenarios, two of them as shown in Fig.~\ref{fig:real-life}.

Our system can give several dynamic information of the prior dynamic object, including its number, average depth, position, pose, velocity, and possible moving direction. The depth value of a person can help the user maintain social distance in a public indoor environment like in a shopping mall. Moreover, when the moving object gets closer to the user and its speed is relatively fast, the reminder of potential risk can be passed to user. The object with high velocity is often dangerous for the visually impaired and this velocity information can enhance current obstacle avoidance modules that mainly use the depth information~\cite{liu2021hida,zhang2021trans4trans}.
We also designed a questionnaire regarding the expected feedback form from our system. Through personal discussion and the results of the online questionnaire, the voice feedback of the system is preferred to be user-related and easily-understandable. The `user-related' indicates that the moving object could affect the user's walking status in short time. For this case, the system should remind users of the potential risk with special signal tone.

\section{Conclusion}
In this work, an assistive system is developed to help people with visual impairments to understand dynamic changes in indoor scenes. 
The static keypoints obtained by sparse-feature visual SLAM are combined with dynamic keypoints, which are obtained by optical flow tracking.
The former aims to estimate the ego-motion robustly and the latter supports the identifying and stable tracking of dynamic objects without additional object models.
However, there are still some limitations of our system. Since the result of panoptic segmentation is hardly perfectly accurate, it leads to some errors when identifying the non-prior dynamic object. Additionally, the computational complexity of this system highly depends on the number of dynamic objects. Some methods of reducing the number of optimized parameters need to be integrated. For the future work, we intend to conduct user experience research, \textit{i.e.}, invite visually impaired volunteers to use our devices and collect feedback, to further improve our system towards more holistic scene perception and reliable navigation assistance.

\bibliographystyle{splncs04}
\bibliography{bibfile}
\end{document}